# Error Analysis for a Navigation Algorithm based on Optical-Flow and a Digital Terrain Map

Oleg Kupervasser, Ronen Lerner, Ehud Rivlin
Dept. of Computer Science, Technion
Haifa 32000, Israel
Hector Rotstein
Rafael - Advanced Defense Systems Ltd.
P.O. Box 2250(39), Haifa 31021, Israel

*Abstract*—The paper deals with the error analysis of a navigation algorithm that uses as input a sequence of images acquired by a moving camera and a Digital Terrain Map (DTM) of the region been imaged by the camera during the motion. The main sources of error are more or less straightforward to identify: camera resolution, structure of the observed terrain and DTM accuracy, field of view and camera trajectory. After characterizing and modeling these error sources in the framework of the CDTM algorithm, a closed form expression for their effect on the pose and motion errors of the camera can be found. The analytic expression provides a priori measurements for the accuracy in terms of the parameters mentioned above.

## I. Introduction

This paper deals with the error analysis of a navigation algorithm that uses as input a sequence of images acquired by a moving camera and a Digital Terrain Map (DTM) of the region been imaged by the camera during the motion. It was shown in a previous work that if the navigation solution (position, velocity and attitude) is approximately known, then the optical flow computed from two image frames can be solved together with the DTM to produce an improved navigation solution. The algorithm was called CDTM since it deals with feature correspondence and a DTM. As opposed to other existing approaches, the algorithm does not require 3D reconstruction or landmark identification. Synthetic and laboratory experiments using a robot and a terrain model were used to demonstrate the algorithm. During the demonstration stage, several parameters of the imaging system and the DTM were assumed, and some of them were also varied for testing the sensitivity of the algorithm. It was clear that if one of the parameters, e.g., the field of view, is severely compromised, then the algorithm did not generate a good quality solution. A more generic tool was desirable to show under which circumstances the CDTM algorithm would produce reasonable results and under which the solution would break down. The error analysis presented in this paper provides this tool. The main sources of error are more or less straightforward to identify:

- Camera resolution. Since the algorithm is based on feature correspondence, the camera resolution affects the accuracy in computing each correspondence constraint.
- Structure of observed terrain and DTM accuracy.

As any other algorithm for terrain navigation, the algorithm requires terrain variability. The results are naturally affected by the accuracy of the available data base.

- Field of view. In order to compute an accurate solution, a sufficiently rich geometry should be available, similar to the GDOP considerations of GNSS navigation. The field-of-view of the camera is critical in this respect.
- Camera trajectory. Since the algorithm is based on the variations between the two consecutive frames under study, these variations should be large enough to provide a good signal-to-noise ration.

One of the main results of the present paper it to establish that after characterizing and modeling the error sources in the framework of the CDTM algorithm, a closed form expression can be found for their effect on the pose and motion errors of the camera. The analytic expression provides *a priori* measurements for the accuracy in terms of the parameters mentioned above. Furthermore, the result was confirmed by using extensive numerical simulations. The main conclusion of this paper is to establish under which generic scenarios the CDTM algorithm as formulated before can be expected to generates accurate estimates for improving a navigation solution.

*A. The CDTM Algorithm*

A vision-based algorithm has been a major research issue during the past decades. Two common approaches for the navigation problem are: *landmarks* and *ego-motion integration*. In the landmarks approach several features are located on the image-plane and matched to their known 3D location. Using the 2D and 3D data the camera's pose can be derived. Few examples for such algorithms are [9], [3]. Once the landmarks were found, the pose derivation is simple and can achieve quite accurate estimates. The main difficulty is the detection of the features and their correct matching to the landmarks set.

In ego-motion integration approach the motion of the camera with respect to itself is estimated. The ego-motion can be derived from the optical-flow field, or from instruments such as accelerometers and gyroscopes. Once the ego-motion was obtained, one can integrate this motion to derive the camera's path. One of the factors that make this approach attractive is that no specific features need to be detected, unlike the previous approach. Several ego-motion estimation algorithms can be found in [1], [13], [2], [5]. The weakness of ego-motion integration comes from the fact that small errors are accumulated during the integration process. Hence, the estimated camera's path is drifted and the pose estimation accuracy decrease along time. If such approach is used it would be desirable to reduce the drift by activating, once in a while, an additional algorithm that estimates the pose directly. In [12], such navigation-system is being suggested. In that work, like in this work, the drift is being corrected using a Digital Terrain Map (DTM). The DTM is a discrete representation of the observed ground's topography. It contains the altitude over the sea level of the terrain for each geographical location. In [12] a patch from the ground was reconstructed using 'structure-from-motion' (SFM) algorithm and was matched to the DTM in order to derive the camera's pose. Using SFM algorithm which does not make any use of the information obtained from the DTM but rather bases its estimate on the flow-field alone, positions their technique under the same critique that applies for SFM algorithms [10].

The algorithm presented in this work does not require an intermediate explicit reconstruction of the 3D world. By combining the DTM information directly with the images information it is claimed that the algorithm is well-conditioned and generates accurate estimates for reasonable scenarios and error sources. In the present work this claim is explored by performing an error analysis on the algorithm outlined above. By assuming appropriate characterization of these error sources, a closed form expression for the uncertainty of the pose and motion of the camera is first developed and then the influence of different factors is studied using extensive numerical simulations.

## II. Problem Definition and Notations

The problem of estimating a navigation solution using image correspondence and a DTM (the C-DTM algorithm) can be briefly described as follows: At any given time instance $t$, a coordinates system $C(t)$ is fixed to a camera in such a way that the $Z$-axis coincides with the optical-axis and the origin coincides with the camera's projection center. At that time instance the camera is located at some geographical location $p(t)$ and has a given orientation $R(t)$ with respect to a global coordinates system $W$ ($p(t)$ is a 3D vector, $R(t)$ is an orthonormal rotation matrix). $p(t)$ and $R(t)$ define the transformation from the camera's frame $C(t)$ to the world's frame $W$, where if $^Cv$ and $^Wv$ are vectors in $C(t)$ and $W$ respectively, then $^Wv = R(t)^Cv + p(t)$.

Consider now two sequential time instances $t_1$ and $t_2$: the transformation from $C(t_1)$ to $C(t_2)$ is given by the translation vector $\Delta p(t_1,t_2)$ and the rotation matrix $\Delta R(t_1,t_2)$, such that $^{C(t_2)}v = \Delta R(t_1,t_2)^{C(t_1)}v + \Delta p(t_1,t_2)$. A rough estimate of the camera's pose at $t_1$ and of the ego-motion between the two time instances – $p_E(t_1)$, $R_E(t_1)$, $\Delta p_E(t_1,t_2)$ and $\Delta R_E(t_1,t_2)$ - are supplied (the subscript letter "E" denotes that this is an estimated quantity).

Also supplied is the optical-flow field: $\{u_i(t_k)\}$ ($i=1...n$, $k=1,2$). For the $i$'th feature, $u_i(t_1) \in IR^2$ and $u_i(t_2) \in IR^2$ represent its locations at the first and second frame respectively.

Using the above notations, the objective of the navigation algorithm is to estimate the true camera's pose and ego-motion: $p(t_1)$, $R(t_1)$, $\Delta p(t_1,t_2)$ and $\Delta R(t_1,t_2)$, using the optical-flow field $\{u_i(t_k)\}$, the DTM and the initial-guess: $p_E(t_1)$, $R_E(t_1)$, $\Delta p_E(t_1,t_2)$ and $\Delta R_E(t_1,t_2)$.

## III. Brief Review of C-DTM

Let $^WG \in IR^3$ be a location of a ground feature point in the 3D world. At two different time instances $t_1$ and $t_2$, this feature point is projected on the image-plane of the camera to the points $u(t_1)$ and $u(t_2)$. Assuming a pinhole model for the camera, then $u(t_1)$, $u(t_2) \in IR^2$. Let $^Cq(t_1)$ and $^Cq(t_2)$ be the homogeneous representations of these locations. As standard, one can think of these vectors as the vectors from the optical-center of the camera to the projection point on the image plane. Using an initial-guess of the pose of the camera at $t_1$, the line passing through $p_E(t_1)$ and $^Cq(t_1)$ can be intersected with the DTM. Any ray-tracing style algorithm can be used for this purpose. The location of this intersection is denoted as $^WG_E$. The subscript letter "E" highlights the fact that this ground-point is the estimated location for the feature point that in general will be different from the true ground-feature location $^WG$. The difference between the true and estimated locations is due to two main sources: the error in the initial guess for the pose and the errors in the determination of $^WG_E$ caused by DTM discretization and intrinsic errors. For a reasonable initial-guess and DTM-related errors, the two points $^WG_E$ and $^WG$ will be close enough so as to allow the linearization of the DTM around $^WG_E$. Let $N$ be the normal of the plane tangent to the DTM at the point $^WG_E$, and define the operators:

$$\mathcal{P}(u,s) \doteq \left(I - \frac{us^T}{s^Tu}\right) \quad (1)$$

$$\mathcal{L} = \frac{q_1 N^T}{N^T R_1 q_1} \quad (2)$$

Then, after some algebra, the single-feature, two-frames C-DTM constraint can be written [7]:

$$\mathcal{P}(q_2, q_2)[p_{12} + R_{12}\mathcal{L}(G_E - p_1)] = 0 \quad (3)$$

This constraint involves the position, orientation and the ego-motion defining the two frames of the camera. Although it involves 3D vectors, it is clear that its rank can not exceed two due to the usage of $P$ which projects $IR^3$ on a two-dimensional subspace. In a numerical implementation it is convenient to use a normalized version of (3):

$$\mathcal{P}(q_2, q_2) \left[ p_{12} + R_{12} \mathcal{L}_i (G_{E_i} - p_1) \right] / |^{C_2}G| = 0 \quad (4)$$

*A. Multiple Features*

The C-DTM constraint can be written for each vector in the optical-flow field. Since overall twelve parameters need to be estimated (six for pose and six for the ego-motion), at least six optical-flow vectors are required for the system solution, although usually more vectors should be used in order to define an over-determined system. Since the constraints are non-linear, an iterative scheme is required to find a solution. A robust algorithm which uses Gauss-Newton iterations and an M-estimator is described in [8]. In the current implementation, a Levenberg-Marquardt method is used whenever Gauss-Newton fails to converge after several iterations. More specifically, suppose that $n$ feature points are tracked in two frames, so that the estimated locations $Q_{Ei}$ and projections onto the image plane $q_{1i}$ and $q_{2i}$ are estimated and measured, respectively, for $i = 1, \cdots, n$. Associated with each $Q_{Ei}$ is the normal vector to the DTM at this point, namely $N_i$. One can then write:

$$\begin{bmatrix} -\mathcal{P}(q_{21}) & \mathcal{P}(q_{21}) \frac{R_{12}q_{11}N_1^T}{N_1^T R_1 q_{11}} \\ -\mathcal{P}(q_{22}) & \mathcal{P}(q_{22}) \frac{R_{12}q_{12}N_2^T}{N_2^T R_1 q_{12}} \\ \vdots & \vdots \\ -\mathcal{P}(q_{2n}) & \mathcal{P}(q_{2n}) \frac{R_{12}q_{1n}N_n^T}{N_n^T R_1 q_{1n}} \end{bmatrix} \begin{bmatrix} p_{12} \\ p_1 \end{bmatrix} =$$

$$\begin{bmatrix} \mathcal{P}(q_{21}) \frac{R_{12}q_{11}N_1^T}{N_1^T R_1 q_{11}} Q_{E1} \\ \mathcal{P}(q_{22}) \frac{R_{12}q_{12}N_2^T}{N_2^T R_1 q_{12}} Q_{E2} \\ \vdots \\ \mathcal{P}(q_{2n}) \frac{R_{12}q_{1n}N_n^T}{N_n^T R_1 q_{1n}} Q_{En} \end{bmatrix} \quad (5)$$

In compact notation:

$$\mathcal{A}_n \begin{bmatrix} p_{12} \\ p_1 \end{bmatrix} = \mathcal{B}_n. \quad (6)$$

Note that $A_n$ and $B_n$ depend on known quantities: the estimated features, the normals of the DTM tangent planes, and the images of the features at the two time instances, together with the unknown orientation $R_1$ and the relative rotation $R_{12}$.

**IV. Error analysis**

In order to evaluate the performance of the algorithm, the objective-function of the minimization process that achieves to meet the C-DTM constraint needs to be defined. For that purpose, a functions $f_i : IR^{12} \to IR^3$ is defined for each feature point $i$, penalizing constraint (4) violation:

$$f_i(p_1, \phi_1, \theta_1, \psi_1, p_{12}, \phi_{12}, \theta_{12}, \psi_{12}) = \mathcal{P}(q_2, q_2) \left[ p_{12} + R_{12} \mathcal{L}_i (G_{E_i} - p_1) \right] / |^{C_2}G| \quad (7)$$

In the above expression, $R_{12}$ and $L_i$ are functions of $(\varphi_{12}, \theta_{12}, \psi_{12})$ and $(\varphi_1, \theta_1, \psi_1)$ respectively. Consider the vector-function $F : IR^{12} \to IR^{3n}$,

$$F(p_1, \phi_1, \theta_1, \psi_1, p_{12}, \phi_{12}, \theta_{12}, \psi_{12}) = [f_1, \ldots, f_n]^T. \quad (8)$$

With this definition, the C-DTM navigation problem has been reduced to one of finding a zero for the function (8). In a practical situation, with $n > 6$ and noisy data, the function will have no zero and hence one will be content to finding a minimum in some sense of the function, for instance, computing the twelve parameters that minimize $M(\theta, D) = ||F(\theta, D)||^2$, where $\theta$ represents the 12-vector of the parameters to be estimated, and $D$ is the concatenation of all the data obtain from the optical-flow and the DTM. Using first order perturbations, it can be shown [4] that the connection between the uncertainty of the data and the uncertainty of the estimated parameters is given by:

$$\Sigma_\theta = \left(\frac{dg}{d\theta}\right)^{-1} \left(\frac{dg}{dD}\right) \Sigma_D \left(\frac{dg}{dD}\right)^T \left(\frac{dg}{d\theta}\right)^{-1} \quad (9)$$

Here, $\Sigma_\theta$ and $\Sigma_D$ represent the covariance matrices of the parameters and the data respectively and the function $g$ is defined as follows:

$$g(\theta, D) \doteq \frac{d}{d\theta} M(\theta, D) = \frac{d}{d\theta} F^T F = 2 J_\theta^T F \quad (10)$$

$J_\theta = dF/du$ is the $(3n \times 12)$ Jacobian matrix of $F$ with respect to the twelve parameters. By ignoring second-order elements, the derivations of $g$ can be approximate by:

$$\frac{dg}{d\theta} \approx 2 J_\theta^T J_\theta \quad (11)$$

$$\frac{dg}{dD} \approx 2 J_\theta^T J_D \quad (12)$$

$J_D = dF/dD$ is defined in a similar way as the $(3n \times m)$ Jacobian matrix of $F$ with respect to the $m$ data components. Assigning (11) and (12) back into (9) yield the following expression:

$$J_T = \left(J_\theta^T J_\theta\right)^{-1} J_\theta^T$$

$$\Sigma_\theta = J_T \cdot \left(J_D \Sigma_D J_D^T\right) \cdot J_T^T \quad (13)$$

The central component $J_D \Sigma_D J_D^T$ represents the uncertainties of $F$ while the pseudo-inverse matrix $(J_\theta^T J_\theta)^{-1} J_\theta^T$ transfers the uncertainties of $F$ to those of the twelve parameters. In the following subsections, $J_\theta, J_D$ and $\Sigma_D$ are explicitly derived.

*A. $J_\theta$ Calculation*

Simple derivations of $f_i$ which is presented in (7) yield the following results:

$$N_P(q_2, {}^{C_2}G) = \mathcal{P}(q_2, q_2) \mathcal{P}({}^{C_2}G, {}^{C_2}G) / |^{C_2}G| \quad (14)$$

$$\frac{df}{dp_1} = -N_P(q_2, {}^{C_2}G) R_{12} \mathcal{L} \quad (15)$$

$$\frac{df}{d\alpha_1} = -N_P(q_2, {}^{c_2}G)R_{12}\mathcal{L}\left(\frac{d}{d\alpha_1}R_1\right)\mathcal{L}(G_E - p_1) \quad (16)$$

$$\frac{df}{dp_{12}} = N_P(q_2, {}^{c_2}G) \quad (17)$$

$$\frac{df}{d\alpha_{12}} = N_P(q_2, {}^{c_2}G)\left(\frac{d}{d\alpha_{12}}R_{12}\right)\mathcal{L}(G_E - p_1) \quad (18)$$

In expressions (16) and (18): $\alpha_1 = \varphi_1, \theta_1, \psi_1$ and: $\alpha_{12} = \varphi_{12}, \theta_{12}, \psi_{12}$. The Jacobian $J_\theta$ is obtained by simple concatenation of the above derivations.

### B. $J_D$ Calculation

Before calculating $J_D$, the data vector $D$ must be explicitly defined. Two types of data are being used by the proposed navigation algorithm: data obtained from the optical-flow field and data obtained form the DTM. Each flow vector starts at $q_1$ and ends at $q_2$. One can consider $q_1$'s location as an arbitrary choice of some ground feature projection, while $q_2$ represent the new projection of the same feature on the second frame. Thus the flow errors are realized through the $q_2$ vectors.

The DTM errors influence the $G_E$ and $N$ vectors in the constraint equation. As before, the DTM linearization assumption will be used. For simplicity the derived orientation of the terrain's local linearization, as expressed by the normal, will be considered as correct while the height of this plane might be erroneous. The connection between the height error and the error of $G_E$ will be derived in the next subsection. Resulting from the above, the $q_1$'s and the $N$'s can be omitted from the data vector $D$. It will be defined as the concatenation of all the $q_2$'s followed by concatenation of the $G_E$'s.

The i'th feature's data vectors: $q_{2i}$ and $G_{Ei}$ appears only in the i'th feature constraint, thus the obtained Jacobian matrix $J_D = [J_q, J_G]$ is a concatenation of two block diagonal matrices: $J_q$ followed by $J_G$. The i'th diagonal block element is the 3x3 matrix $df_i/dq_{2i}$ and $df_i/dG_{Ei}$ for $J_q$ and $J_G$ respectively:

$$\frac{df}{dq_2} =$$

$$\frac{-1}{\|q_2\|^2}\left[(q_2^T \cdot {}^{c_2}G)\,\mathrm{I} + q_2 \cdot {}^{c_2}G^T\right]\mathcal{P}(q_2, q_2)/|{}^{c_2}G| \quad (19)$$

$$\frac{df}{dG_E} = N_P(q_2, {}^{c_2}G)R_{12}\mathcal{L} \quad (20)$$

${}^{c_2}G$ in expression (19) is the ground feature $G$ in the second camera frame.

\
### C. $\Sigma_D$ Calculation

As mention above, the data-vector D is constructed from concatenation of all the $q_2$'s followed by concatenation of the $G_E$'s. Thus $\Sigma_D$ should represent the uncertainty of these elements. Since the $q_2$'s and the $G_E$'s are obtained from two different and uncorrelated processed the covariance relating them will be zero, which leads to a two block diagonal matrix:

$$\Sigma_D = \begin{bmatrix} \Sigma_q & 0 \\ 0 & \Sigma_G \end{bmatrix} \quad (21)$$

In this work the errors of image locations and DTM height are assumed to be additive zero-mean Gaussian distributed with standard-deviation of $\sigma_I$ and $\sigma_h$ respectively. Each $q_2$ vector is a projection on the image plane where a unit focal-length is assumes. Hence, there is no uncertainty about its $z$-component. Since a normal isotropic distribution was assumed for the sake of simplicity, the covariance matrix of the image measurements is defined to be:

$$\Sigma_{q_i} = \sigma_I^2 \cdot \begin{bmatrix} 1 & & \\ & 1 & \\ & & 0 \end{bmatrix} \quad (22)$$

and $\Sigma_q$ is the matrix with the $\Sigma_{qi}$'s along its diagonal.

In [11] the accuracy of location's height obtained by interpolation of the neighboring DTM grid points is studied. The dependence between this accuracy and the specific required location, for which height is being interpolated, was found to be negligible. Here, the above finding was adopted and a constant standard-deviation was set to all DTM heights measurements. Although there is dependence between close $G_E$'s uncertainties, this dependence will be ignored in the following derivations for the sake of simplicity. Thus, a block diagonal matrix is obtained for $\Sigma_G$ containing the 3x3 covariance matrices $\Sigma_{Gi}$ along its diagonal which will be derived as follows: consider the ray sent from $p_1$ along the direction of $R_1q_1$. This ray should have intersected the terrain at $G_E = p_1 + \lambda R_1 q_1$ for some $\lambda$, but due to the DTM height error the point $\tilde{G}_E = \left(\tilde{x}, \tilde{y}, \tilde{h}\right)^T$ was obtained. Let $h$ be the true height of the terrain above $(\tilde{x}, \tilde{y})$ and $H = (\tilde{x}, \tilde{y}, h)$ be the 3D point on the terrain above that location.

Using that H belongs to the true terrain plane one obtains:

$$N^T(G_E - H) = N^T(p_1 + \lambda R_1 q_1 - H) = 0 \quad (23)$$

Extracting $\lambda$ from (23) and assigning it back to $G_E$'s expression yields:

$$G_E = p_1 + R_1\mathcal{L}(H - p_1) \quad (24)$$

For $G_E$'s uncertainty calculation the derivative of $G_E$ with respect to $h$ should be found:

$$\frac{dG_E}{dh} = R_1 \mathcal{L} \cdot \begin{pmatrix} 0 & 0 & 1 \end{pmatrix}^T = \frac{R_1 q_1}{N^T R_1 q_1} \quad (25)$$

The above result was obtained using the fact that the z-component of $N$ is 1: $N = \begin{pmatrix} -\nabla D\overset{\smile}{T}M & 1 \end{pmatrix}^T$. Finally, the uncertainty of $G_E$ is expressed by the following covariance-matrix:

$$\Sigma_{G_i} = \left(\frac{dG_E}{dh}\right) \cdot \sigma_h^2 \cdot \left(\frac{dG_E}{dh}\right)^T = \sigma_h^2 \cdot \frac{R_1 q_1 q_1^T R_1^T}{(N^T R_1 q_1)^2} \quad (26)$$

D. $\sum_{C_2}$ Calculation

The algorithm presented in this work estimates the pose of the first camera frame and the ego-motion. Usually, the most interesting parameters for navigation purpose will be the second camera frame since it reflects the most updated information about the platform location. The second pose can be obtained in a straightforward manner as the composition of the first frame pose together with the camera ego-motion:

$$p_2 = p_1 - R_1 R_{12}^T p_{12} \quad (27)$$
$$R_2 = R_1 R_{12}^T \quad (28)$$

The uncertainty of the second pose estimates will be described by a 6×6 covariance matrix that can be derived from the already obtained 12×12 covariance matrix $\sum_\theta$ by multiplication from both sides with $J_{C2}$. The last notation is the Jacobian of the six $C_2$ parameters with respect to the twelve parameters mentioned above. For this purpose, the three Euler angles $\varphi_2$, $\theta_2$ and $\psi_2$ need to be extracted from (28) using the following equations:

$$\phi_2 = \arctan\left(\frac{R_2(2,3)}{R_2(3,3)}\right) \quad (29)$$
$$\theta_2 = \arcsin(-R_2(1,3)) \quad (30)$$
$$\psi_2 = \arctan\left(\frac{R_2(1,2)}{R_2(1,1)}\right) \quad (31)$$

Simple derivations and concatenation of the above expressions yields the required Jacobian which is used to propagate the uncertainty from $C_1$ and the ego-motion to $C_2$. The resulting covariance matrix $\sum_{C2}$ is the same as the measurement covariance matrix $R_k$ that can be used for formulating a Kalman filter [6].

$$R_k = \Sigma_{C_2} \quad (32)$$

**V. Existence of a solution**

The error analysis in the previous section implicitly assumes the existence of a solution to the minimization problem which is close to the actual navigation solution. It this assumptions does not hold, the minimization procedure may not converge or may get stack on a local minimum far from the true solutions. This kind of undesirable behavior may appear under one of the following circumstances:

1) The constraints include one or more outliers that dominate the overall solution.
2) The configuration of the feature points gives rise to a degenerate under-constrained system of equations in such a way that the position or orientation errors may become undefined. This pathological case may result, from instance, from using a small number of feature points, observing flat terrain with no variations or using a camera with relatively small field of view.
3) The initial position and orientation employed in the iterations process are far from the true values and the nonlinear optimization procedure fails to converge to a true solution. The purpose of the next subsections is to consider some threshold conditions that will guarantee the avoidance of pathological situations.

*A. Dealing with Outliers*

Outliers may be roughly classified in three classes:
1) Outliers due to incorrect feature matching between frames.
2) Outliers caused by the terrain shape, and
3) Outliers due to DTM/terrain mismatch.

Fig.1 shows the effect of the latter two classes. The outliers caused by the terrain shape appear for terrain features located close to large depth variations. For example, consider two hills, one closer to the camera, the other farther away, and a terrain feature $Q$ located on the closer hill. The ray-tracing algorithm using the erroneous pose may "miss" the proximal hill and erroneously places the feature on the distal one. Needless to say, the error between the true and estimated locations is not covered by the linearization. To visualize the errors introduced by a relatively large DTM-actual terrain mismatch, suppose a building was present on the terrain when the DTM was acquired, but is no longer there when the experiment takes place. The ray-tracing algorithm will locate the feature on the building although the true terrain-feature belongs to a background that is now visible. As discussed above, the multi-feature constraint is solved in a least-squares sense for the pose and motion variables. Given the sensitivity of least-squares to incorrect data, the inclusion of one or more outliers may result in the convergence to a wrong solution. A possible way to circumvent this difficulty that appears to work well in practice is to use an M-estimator as proposed in [7]

*B. Degenerate Configurations*

When feature points and/or the DTM are in a degenerate configuration, the matrix $J_\theta^T W J_\theta$ becomes singular, suggesting that one can detect critical configurations by tracking the condition number or other measurement of how close this matrix is to singular. Similarly, the covariance matrices $\sum_{C2}$ and $\sum_\theta$ become singular or close to singular when the number of feature points is small, the terrain has no features that can be used for differentiation or the camera FOV is small. Measuring how close these matrices are to singular and threshold can be used to prevent the navigation update from computing an incorrect navigation solution.

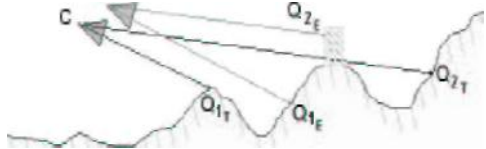

Fig. 1. Two classes of outliers

## C. Large Initial Errors

The linearization algorithm may fail if the initial navigation solution is too far from the true one in such a way that the approximation of the DTM by a local plane on the assumed location of the feature point does not hold. It is then important to define a measurement that will prevent incorrect computations and quality assessments. Let $P_k^-$ be a covariance matrix that measures the priori uncertainty on the navigation solution. Then a measurement reject policy may be formulated using this covariance and the state innovation [14].

## VI. Simulations Results

The purpose of this section is to study the influence of the different factors considered above on the accuracy of the C-DTM algorithm under a variety of simulated scenarios. Each tested scenario is characterized by the following parameters: number of optical-flow features being used by the algorithm, image resolution, the grid spacing of the DTM (also referred as DTM resolution), the amplitude of hills/mountains on the observed terrain, and the magnitude of the ego-motion. On each simulation run, all parameters except the examined one are taken from a predefined parameters set. In this *nominal scenario,* a camera with 400x400 image resolution moves at a constant altitude of 500 m. The terrain model dimensions are 3x3 km with 300 m elevation differences (Fig.7(b)). A 30m DTM grid is used to model the terrain (Fig.4(c)). The DTM resolution leads to a standard-deviation of about 2.4m for the height measurements. The default-scenario also defines the number of optical-flow features to about 170, with an ego-motion of $\|p_{12}\| = 40m$ and $\|(\varphi_{12}, \theta_{12}, \psi_{12})\| = 10°$ between two images. Since parameters are varied one at a time, the results summarized next may be considered a *sensitivity* study, where different):

values were examined by performing 150 random tests for each tested value.

Fig.2 summarizes the effect of the number of optical-flow features on the accuracy of rotation and the ego-motion recovery. Fig.2(a) presents the standard-deviations at the second frame of the camera while the deviations of the ego-motion are shown in Fig.2(b). As expected, the accuracy improves as the number of features increases, although the improvement saturates at about 150 features.

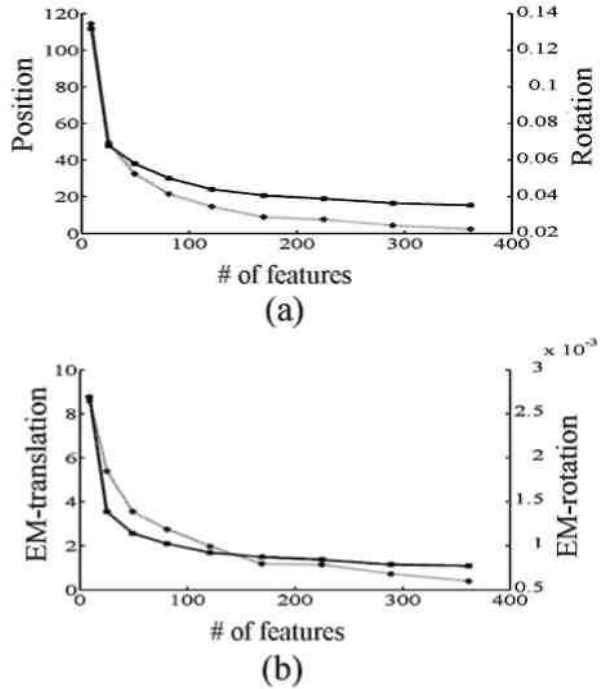

Fig. 2. Number of feature points vs. (a) position and orientation STD at second frame, and (b) recovered translation and rotation error.

Fig.3) is the result of the study of the effect of image resolution. In order to perform this study, it was assumed that registration could achieve half-pixel accuracy, with the size of the pixels dictated by the assumed image resolution. As expected, accuracy improves as image resolution increases due to increase quality in the optical-flow data.

Next, the effect of different DTM grid spacing was investigated, with grid size varying from 10 to 190 m. Results are summarized in Fig.4 showing that accuracy appears to be inversely proportional to grid-spacing, given that resolution affects the height uncertainty and therefore overall accuracy. Indeed, Fig.6 shows that the standard-deviation of the DTM heights increases linearly with respect to the DTM grid spacing.

The next study addresses the importance of terrain variations. In is intuitively clear (and follows directly from the C-DTM constraint) that when flying above a planar terrain ground features do not contain the required information for computing the camera pose so that constraints become singular. As the variability of the terrain increases,

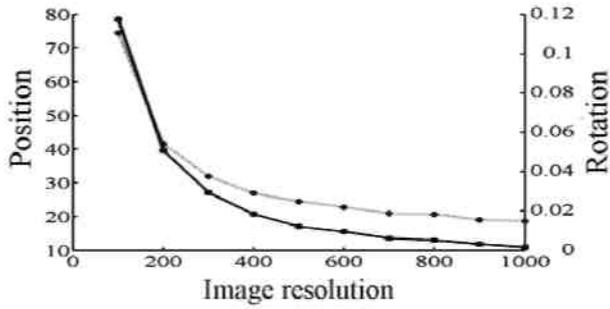

(a)

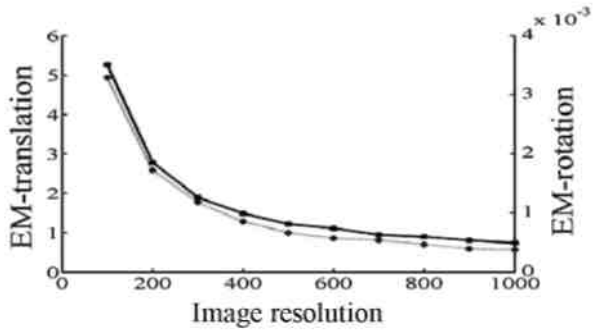

(b)

Fig. 3. Image resolution vs. (a) position and orientation standard deviation at second frame and (b) recovered translation and rotation.

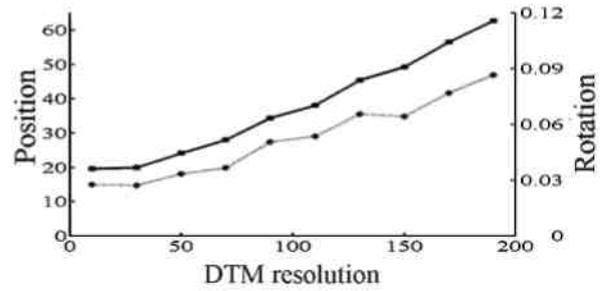

(a)

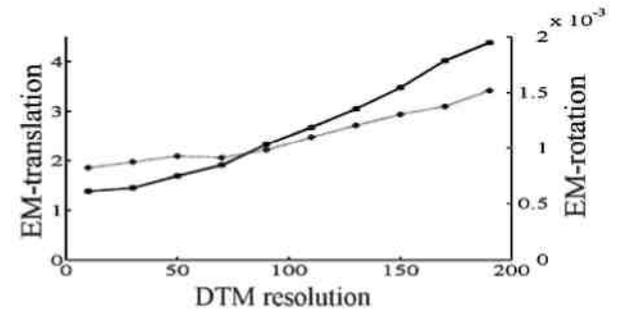

(b)

Fig. 5. Grid spacing vs. (a) position and orientation standard-deviation at second frame and (b) recovered translation and rotation.

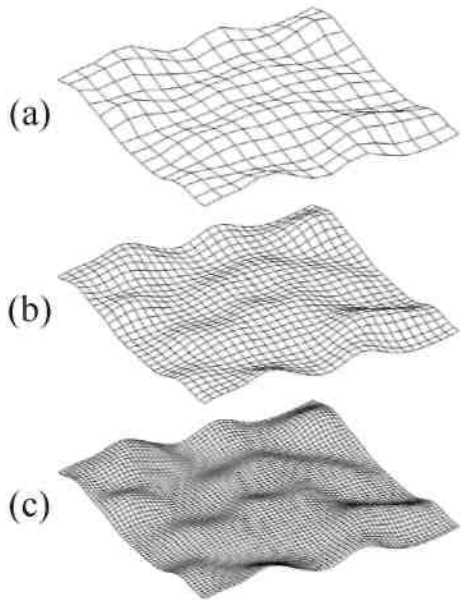

Fig. 4. (a) Grid spacing = 190m. (b) Grid spacing = 100m. (c) Grid spacing = 30m

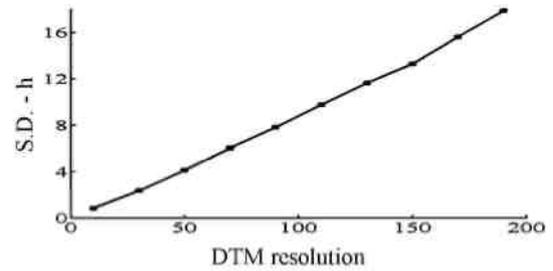

Fig. 6. Grid-spacing vs. DTM height error.

the features become more informative and better estimates are obtained. In the study, the DTM elevation variations were scaled to vary from 50 to 450 m (Fig.7). It is worth mentioning that the terrain structure plays a crucial role in the camera pose estimation and the translational component, while it has no direct affect on the ego-motion rotational component. This is not surprising since C-DTM should reduce, at worst, to the epipolar constraint. Results are summarized in Fig.8.

As in all other methods for recovering motion from a sequence of images, the translation baseline between two frames has a critical impact in the accuracy of the C-DTM method. The final study attempts to quantify this effect by varying the baseline from 5 to 95 m. Results are summarized in Fig.9.

*A. Numerical simulation*

Up to this point, error analysis and assessment was performed under the assumption that the C-DTM is a standalone method for computing one-shot navigation solutions. More generally, C-DTM can be used as a soft-sensor for updating a navigation filter and hence the effect of the errors should be considered in terms of overall navigation performance. In [7] a Kalman Filter was proposed a synthetic DTM was constructed by taking a real ground model as basic cell, and multiplying it until it includes all the region of the flight (Fig. 12). An IMU error model was also included.

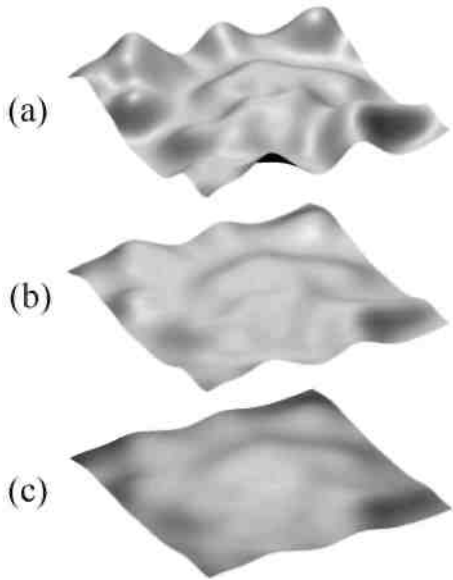

Fig. 7. Different DTM elevation scalings: (a) 150 m, (b) 300 m, (c) 450 m.

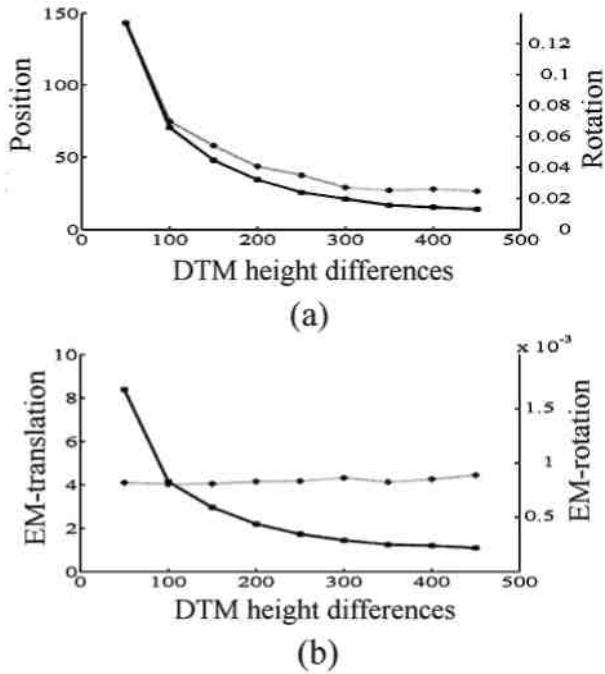

Fig. 8. Terrain variability vs. (a) position and orientation standard deviation at second frame and (b) recovered translation and rotatio

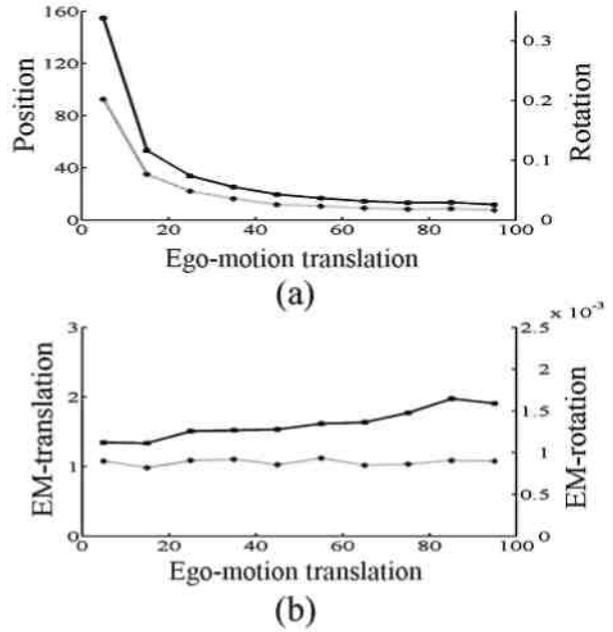

Fig. 9. Translation baseline vs. (a) position and orientation standard-deviation at second frame and (b) recovered translation and rotation.

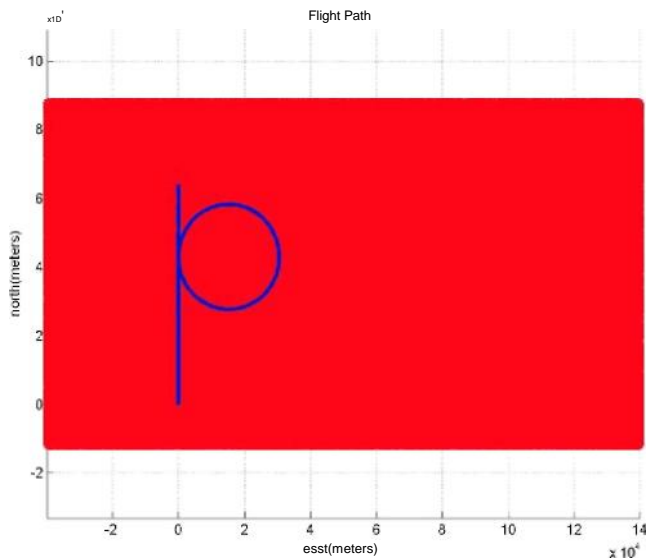

Fig. 10. Flight ground-trajectory.

for fusing inertial navigation and Inertial navigatio systems (INS) with Optical-Flow and a Digital Terrain map error. The purpose of this section is to evaluate the effects of the different error sources on the resulting navigation scheme. For this purpose, the ground trajectory illustrated in (Fig.10) was flown at a constant velocity of 200 m/sec. for three different altitudes: 700, 1000 and 3000 m. To simplify the simulations and memory management,

Remaining camera and simulation parameters were selected as follows:
- Camera FOV: 60 degree
- Number of feature per frame pair: 100 and 120.

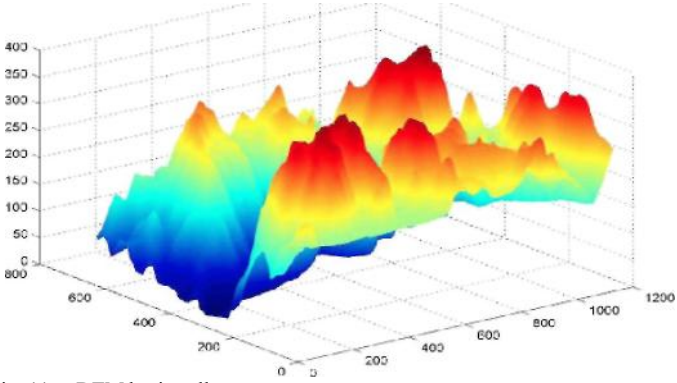

Fig. 11. DTM basic cell.

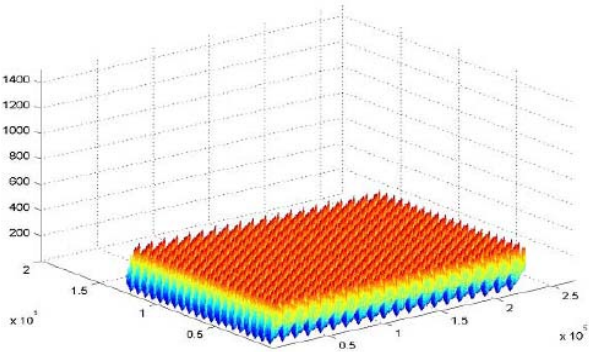

Fig. 12. DTM resulting from basic cell cloning.

- Camera resolutions: 500x500, 1000x1000 and 4000x4000.
- No outliers.
- Translation baseline between two frames: 30, 50 and 200 m.
- Time interval between two frames: 5, 15 and 30 sec.

| Flight height | 700m | 1000m | 3000m |
|---|---|---|---|
| Free | 900m | 130m | 1300 m |
| Compensated-x | 25 m | 20 m | 100 m |
| Free-y | 1000m | 2000m | 4000 m |
| Compensated-y | 25 m | 20 m | 100 m |
| Free-z | 250m | 180m | 250 m |
| Compensated-z | 25m | 20m | 150m |

Table 1. Maximum errors vs. height.

To illustrate the results, Tables 1 and 2 contain the errors for two of the parameters: flight altitude and camera

| Resolution | 500x500 | 1000x1000 | 4000x4000 |
|---|---|---|---|
| Max x error | 50m | 20m | 10m |
| Max y error | 50m | 20m | 10m |
| Max z error | 35m | 20m | 10m |

Table 2. Effect of resolution on max. error.

## VII. Conclusion and Future Work

In this paper an error study was conducted on the C-DTM approach for computing a navigation solution based on correspondence or optical flow and a digital terrain map. A A linear approximation on the C-DTM constraint was performed in order to quantify the effect of various inaccuracies on the estimated navigation solution. Furthermore, situations under which the solution to the C-DTM constraint may lead to a wrong fix were analyzed together with computable indicators. Results were illustrated using a sensitivity study of the method under different testing and simulation conditions. It was shown that the algorithm behaved robustly even with relatively noisy data and a challenging environment. Following the analysis, it can be argued that the proposed algorithm can be effectively used as part of a navigation system of autonomous flying vehicles. Specific conclusions are as follows:

1) The most critical sensitivity parameter is the FOV of the camera used for constructing the C-DTM constraint. Results are excellent for FOV=$60^o$ but fail to converge for FOV<$8^o$ when the geometry of the constraints become degenerate.
2) C-DTM is also sensitive to small translational baseline or when the observed ground patch is too small and with little terrain variations.
3) Camera resolution also affects the accuracy of the method.
4) Flight altitude has two different effects. For relatively low altitudes, accuracy increases as a function of height since the size of the ground patch increases and consequently larger translation baselines can be used. At some point and for a fixed camera resolution, accuracy begins to deteriorate due to decreased ground resolution.

The results obtained by using the C-DTM constraint can be improved by modifying some of the implementation aspects of the solution. From an image processing viewpoint, it is possible to use more structured features or some known geographical entities like valleys and hills occluding boundaries. From a numerical perspective, one can consider to improve the initial solution before starting the computations by estimating $R_{12}$ and $p_{12}$ (up to a constant) by using the more standard epipolar constraint, and only then using a simplified version of the C-DTM to estimate $R_1$ and actual position.

More ambitiously, current research work is focused on reducing the impact of some of the vision-related parameters on an overall navigation solution. For example, if the pure C-DTM solution is challenging due to a relatively small field of view, low altitude or short baseline, information can be integrated using a navigation filter (e.g., a Kalman filter) to work iteratively over several frames. Some progress has already been achieved by considering the inclusion of the C-DTM constraint directly on the navigation filter.

# References

[1] J. L. Barron and R. Eagleson. Recursive estimation of time-varying motion and structure parameters. *Patt. Recognition*, 29(5):797–818, 1996.

[2] A. Chiuso, P. Favaro, H. Jin, and S. Soatto. MFm: 3-D motion from 2-D motion causally integrated over time. In *Proc. of the European Conf. of Comp. Vision*, 2000.

[3] P. David, D. DeMenthon, and R. Duraiswami. Simultaneous Pose and Correspondence Determination Using Line Features. In *cvpr*, 2003.

[4] R. M. Haralick. Propagating covariances in computer vision. In *icpr*, pages 493–498, Jerusalem, Israel, September 1994.

[5] M. Irani, B. Rousso, and S. Peleg. Robust recovery of ego-motion. In *Proc. Of Comp. Analysis of Images and Patt.*, pages 371–378, 1993.

[6] O. Kupervasser, E. Rivlin, and H. Rotstein. A navigation filter for fusing dtm/correspondence updates. In preparation, 2008.

[7] R. Lerner, E.Rivlin, and H. P. Rotstein. Pose and motion recovery from correspondence and a digital terrain map. *IEEE Trans. on Patt. Analysis and Machine Intelligence*, 28(9):1404–1417, 2006.

[8] R. Lerner, H. Rotstein, and E. Rivlin. Error analysis of an algorithm for pose and motion recovery from correspondence and a digital terrain map. In preparation, 2007.

[9] Y. Liu and M. A. Rodrigues. Statistical image analysis for pose estimation without point correspondences. *Pattern Recognition Letters*, 22:1191–1206, 2001.

[10] J. Oliensis. A critique of structure-from-motion algorithms. *Comp. Vision and Image Understanding*, 80:172–214, 2000.

[11] W. G. Rees. The accuracy of digital elevation models interpolated to higher resolutions. *ijrs*, 21(1):7–20, 2000.

[12] D. G. Sim, R. H. Park, R. C. Kim, S. U. Lee, and I. C. Kim. Integrated position estimation using aerial image sequences. *IEEE Trans. on Patt. Analysis and Machine Intelligence*, 24(1):1–18, 2002.

[13] T. Tian, C. Tomasi, and D. Heeger. Comparison of approaches to egomotion computation. *Proc. IEEE Conf. Comp. Vision Patt. Recog.*, pages 315–320, 1996.

[14] X. R. L. Y. Bar-Shalom and T. Kirubarajan. *Estimation with Applications to Tracking and Navigation*. John Wiley & Sons Inc., 2001.